%% file: main.tex
\documentclass[10pt,twocolumn,letterpaper]{article}

\usepackage{cvpr}
\usepackage{times}
\usepackage{epsfig}
\usepackage{graphicx}
\usepackage{amsmath}
\usepackage{amssymb}
\usepackage{float}

\newcommand{\model}{VectorNet\xspace}

\newcommand{\eat}[1]{}

\usepackage[breaklinks=true,bookmarks=false]{hyperref}

\cvprfinalcopy % *** Uncomment this line for the final submission

\def\cvprPaperID{4284} % *** Enter the CVPR Paper ID here

\ifcvprfinal\fi
\begin{document}

%%%%%%%%% TITLE
\title{VectorNet: Encoding HD Maps and Agent Dynamics from \\Vectorized Representation}

\author{Jiyang Gao$^{1*}$ \quad Chen Sun$^{2*}$ \quad Hang Zhao$^1$  \quad Yi Shen$^1$ \quad  \\  Dragomir Anguelov $^1$  \quad Congcong Li$^1$ \quad Cordelia Schmid $^2$ \\
$^1$Waymo LLC \qquad $^2$ Google Research \\
{\tt\small \{jiyanggao, hangz, yshen, dragomir, congcongli\}@waymo.com,\quad \{chensun, cordelias\}@google.com}}

\maketitle
\thispagestyle{empty}
\newcommand\blfootnote[1]{%
  \begingroup
  \renewcommand\thefootnote{}\footnote{#1}%
  \addtocounter{footnote}{-1}%
  \endgroup
}

\blfootnote{$*$ equal contribution.}

%%%%%%%%% ABSTRACT
\input{sections/abstract}

\input{sections/intro}
\input{sections/related}

\input{sections/method}

\input{sections/experiments}

\input{sections/conclusion}

\clearpage

{\small
\bibliographystyle{ieee_fullname}
\bibliography{egbib}
}

\end{document}

%% file: sections/abstract.tex
\begin{abstract}
Behavior prediction in dynamic, multi-agent systems is an important problem in the context of self-driving cars, due to the complex representations and interactions of road components, including moving agents (\eg pedestrians and vehicles) and road context information (\eg lanes, traffic lights). This paper introduces \model, a hierarchical graph neural network that first exploits the spatial locality of individual road components represented by vectors and then models the high-order interactions among all components. In contrast to  most recent approaches, which render trajectories of moving agents and road context information as bird-eye images and encode them with convolutional neural networks (ConvNets), our approach operates on a  vector representation. By operating on the vectorized high definition (HD) maps and agent trajectories, we avoid lossy rendering and computationally intensive ConvNet encoding steps. To further boost VectorNet's capability in learning context features, we propose a novel auxiliary task to recover the randomly masked out map entities and agent trajectories based on their context. We evaluate \model on our in-house behavior prediction benchmark and the recently released Argoverse forecasting dataset. Our method achieves on par or better performance than the competitive rendering approach on both benchmarks while saving over 70\% of the model parameters with an order of magnitude reduction in FLOPs. It also outperforms the state of the art on the Argoverse dataset.
\end{abstract}

%% file: sections/intro.tex
\section{Introduction}

This paper focuses on behavior prediction in complex multi-agent systems, such as self-driving vehicles. The core interest is to find a unified representation which integrates the agent dynamics, acquired by perception systems such as object detection and tracking, with the scene context, provided as prior knowledge often in the form of High Definition (HD) maps. Our goal is to build a system which learns to predict the intent of vehicles, which are parameterized as trajectories.

\begin{figure}
    \centering
    \includegraphics[width=\linewidth]{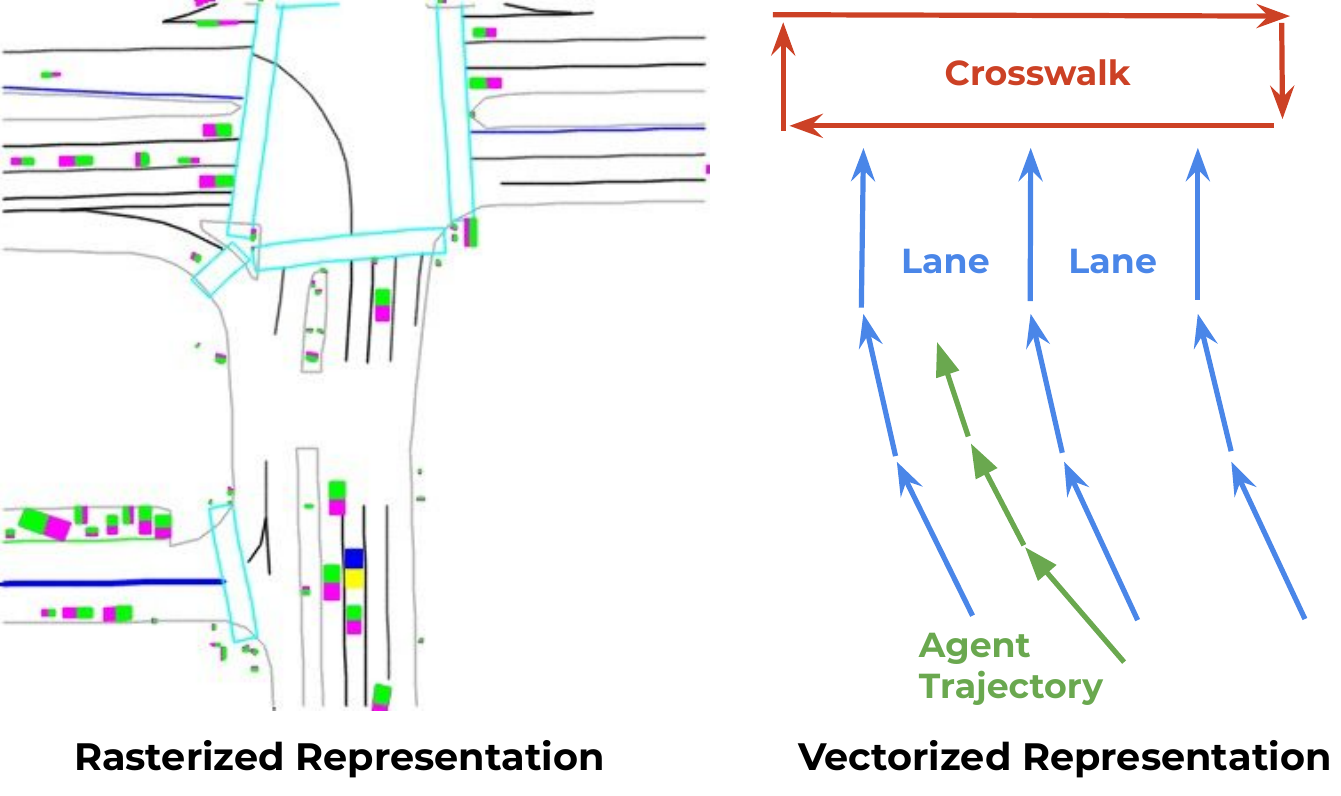}
    \caption{Illustration of the rasterized rendering (left) and vectorized approach (right) to represent high-definition map and agent trajectories.}
    \label{fig:vector}
\end{figure}

Traditional methods for behavior prediction are rule-based, where multiple behavior hypotheses are generated based on constraints from the road maps. More recently, many learning-based approaches are proposed~\cite{casas2018intentnet, chai2019multipath, cui2019multimodal, hong2019rules}; they offer the benefit of having probabilistic interpretations of different behavior hypotheses, but require building a representation to encode the map and trajectory information. Interestingly, while the HD maps are highly structured, organized as entities with location (\eg lanes) and attributes (\eg a green traffic light), most of these approaches choose to render the HD maps as color-coded attributes (Figure~\ref{fig:vector}, left), which requires manual specifications; and encode the scene context information with ConvNets, which have limited receptive fields. This raise the question: can we learn a meaningful context representation directly from the structured HD maps?

%Despite their promising results, the additional steps of rendering and deep ConvNets makes them less desirable for self-driving application, where a low latency behavior predictor is desired. This raises the question: can we learn meaningful context representation directly from the structured HD maps?

\begin{figure*}[htp!]
    \centering
    \includegraphics[width=0.95\linewidth]{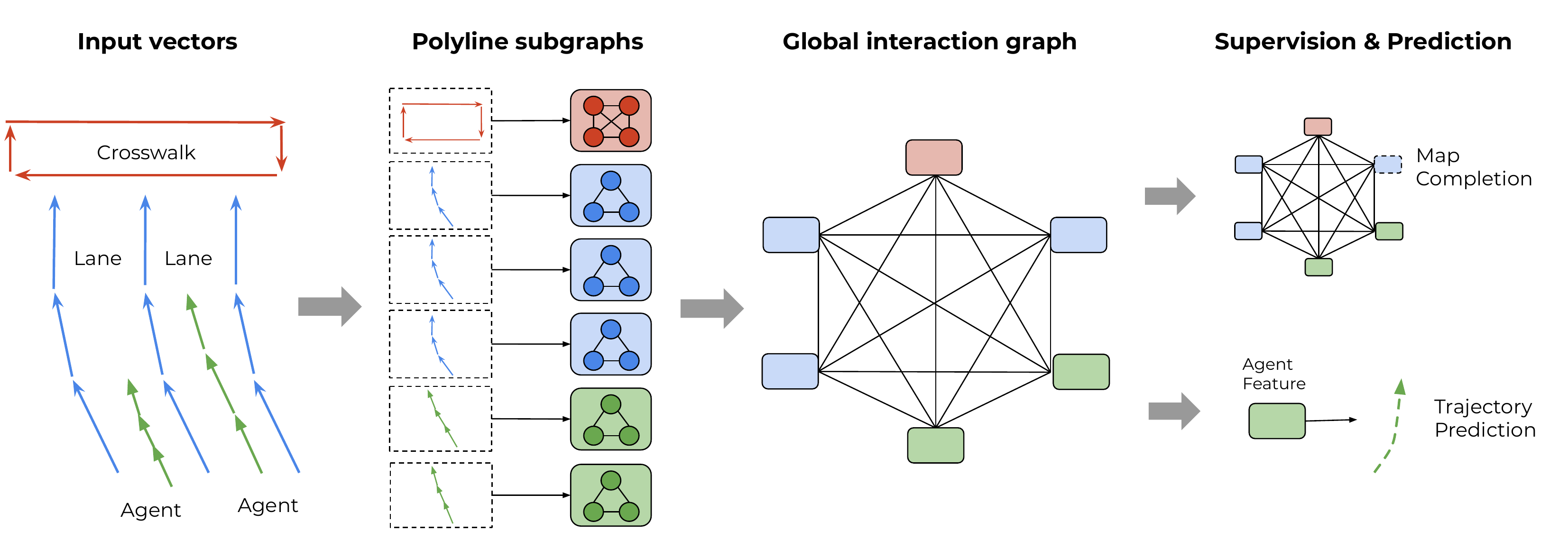}
    \caption{An overview of our proposed \model. Observed agent trajectories and map features are represented as sequence of vectors, and passed to a local graph network to obtain polyline-level features. Such features are then passed to a fully-connected graph to model the higher-order interactions. We compute two types of losses: predicting future trajectories from the node features corresponding to the moving agents and predicting the node features when their features are masked out.}
    \label{fig:flowchart}
    %\vspace{-0.9em}
\end{figure*}

We propose to learn a unified representation for multi-agent dynamics and structured scene context directly from their vectorized form (Figure~\ref{fig:vector}, right). The geographic extent of the road features can be a point, a polygon, or a curve in geographic coordinates. For example, a lane boundary contains multiple control points that build a spline; a crosswalk is a polygon defined by several points; a stop sign is represented by a single point. All these geographic entities can be closely approximated as polylines defined by multiple control points, along with their attributes. Similarly, the dynamics of moving agents can also be approximated by polylines based on their motion trajectories. All these polylines can then be represented as sets of vectors.

We use graph neural networks (GNNs) to incorporate these sets of vectors. We treat each vector as a node in the graph, and set the node features to be the start location and end location of each vector, along with other attributes such as polyline group id and semantic labels. The context information from HD maps, along with the trajectories of other moving agents are propagated to the target agent node through the GNN. We can then take the output node feature corresponding to the target agent to decode its future trajectories.

Specifically, to learn competitive representations with GNNs, we observe that it is important to constrain the connectivities of the graph based on the spatial and semantic proximity of the nodes. We therefore propose a hierarchical graph architecture, where the vectors belonging to the same polylines with the same semantic labels are connected and embedded into polyline features, and all polylines are then fully connected with each other to exchange information. We implement the local graphs with multi-layer perceptrons, and the global graphs with self-attention~\cite{vaswani2017attention}. An overview of our approach is shown in Figure~\ref{fig:flowchart}.

Finally, motivated by the recent success of self-supervised learning from sequential linguistic~\cite{devlin2018bert} and visual data~\cite{Sun_2019_arXiv}, we propose an auxiliary graph completion objective in addition to the behavior prediction objective. More specifically, we randomly mask out the input node features belonging to either scene context or agent trajectories, and ask the model to reconstruct the masked features. The intuition is to encourage the graph networks to better capture the interactions between agent dynamics and scene context. In summary, our contributions are:
\begin{itemize}
\setlength\itemsep{-0.4em}
    \item We are the first to demonstrate how to directly incorporate vectorized scene context and agent dynamics information for behavior prediction.
    \item We propose the hierarchical graph network \model and the node completion auxiliary task.
    \item We evaluate the proposed method on our in-house behavior prediction dataset and the Argoverse  dataset, and show that our method achieves on par or better performance over a competitive rendering baseline with 70\% model size saving and an order of magnitude reduction in FLOPs. Our method also achieves the state-of-the-art performance on Argoverse.
\end{itemize}

%% file: sections/related.tex
\section{Related work}

\noindent\textbf{Behavior prediction for autonomous driving.} Behavior prediction for moving agents has become increasingly important for autonomous driving applications~\cite{chang2019argoverse,NGSIM,highDdataset}, and high-fidelity maps have been widely used to provide context information. For example, IntentNet~\cite{casas2018intentnet} proposes to jointly detect vehicles and predict their trajectories from LiDAR points and rendered HD maps. Hong \etal~\cite{hong2019rules} assumes that vehicle detections are provided and focuses on behavior prediction by encoding entity interactions with ConvNets. Similarly, MultiPath~\cite{chai2019multipath} also uses ConvNets as encoder, but adopts pre-defined trajectory anchors to regress multiple possible future trajectories. PRECOG~\cite{Rhinehart_2019_ICCV} attempts to capture the future stochasiticity by flow-based generative models. Similar to~\cite{chai2019multipath,hong2019rules,Rhinehart_2019_ICCV}, we also assume the agent detections to be provided by an existing perception algorithm. However, unlike these methods which all use ConvNets to encode rendered road maps, we propose to directly encode vectorized scene context and agent dynamics.

\noindent\textbf{Forecasting multi-agent interactions.} Beyond the autonomous driving domain, there is more general interest to predict the intents of interacting agents, such as for pedestrians~\cite{SocialLSTM,SocialGAN,robicquet2016learning}, human activities~\cite{RAF} or for sports players~\cite{felsen17,sun2019stochastic,Yeh_2019_CVPR,ZhanBasketball}. In particular, Social LSTM~\cite{SocialLSTM} models the trajectories of individual agents as separate LSTM networks, and aggregates the LSTM hidden states based on spatial proximity of the agents to model their interactions. Social GAN~\cite{SocialGAN} simplifies the interaction module and proposes an adversarial discriminator to predict diverse futures. Sun \etal~\cite{sun2019stochastic} combines graph networks~\cite{Battaglia2018arxiv} with variational RNNs~\cite{VRNN} to model diverse interactions. The social interactions can also be inferred from data: Kipf \etal~\cite{Kipf2018icml} treats such interactions as latent variables; and graph attention networks~\cite{VAIN,GAT} apply self-attention mechanism to weight the edges in a pre-defined graph. Our method goes one step further by proposing a unified hierarchical graph network to jointly model the interactions of multiple agents, and their interactions with the entities from road maps.

\noindent\textbf{Representation learning for sets of entities.} Traditionally machine perception algorithms have been focusing on high-dimensional continuous signals, such as images, videos or audios. One exception is 3D perception, where the inputs are usually in the form of unordered point sets, given by depth sensors. For example, Qi \etal propose the PointNet model~\cite{qi2017pointnet} and PointNet++~\cite{qi2017pointnet++} to apply permutation invariant operations (\eg max pooling) on learned point embeddings. Unlike point sets, entities on HD maps and agent trajectories form closed shapes or are directed, and they may also be associated with attribute information. We therefore propose to keep such information by vectorizing the inputs, and encode the attributes as node features in a graph.

\noindent\textbf{Self-supervised context modeling.} Recently, many works in the NLP domain have proposed modeling language context in a self-supervised fashion~\cite{devlin2018bert,gpt2}. Their learned representations achieve significant performance improvement when transferred to downstream tasks. Inspired by these methods, we propose an auxiliary loss for graph representations, which learns to predict the missing node features from its neighbors. The goal is to incentivize the model to better capture interactions among nodes.

%% file: sections/method.tex
\section{VectorNet approach}

This section introduces our \model approach. We first describe how to vectorize agent trajectories and HD maps. Next we present the hierarchical graph network which aggregates local information from individual polylines and then globally over all trajectories and map features. This graph can then be used for behavior prediction.
%A system overview can be found in Figure~\ref{fig:flowchart}.

\subsection {Representing trajectories and maps}
%\HZ{Given the name of VectorNet, we can probably start with something big here: VectorNet is a general geometric encoding technique: for points, lines, curves, meshes. It could find applications in many graphics problems.}

Most of the annotations from an HD map are in the form of splines (\eg lanes), closed shape (\eg regions of intersections) and points (\eg traffic lights), with additional attribute information such as the semantic labels of the annotations and their current states (\eg color of the traffic light, speed limit of the road). For agents, their trajectories are in the form of directed splines with respect to time. All of these elements can be approximated as sequences of vectors: for map features, we pick a starting point and direction, uniformly sample key points from the splines at the same spatial distance, and sequentially connect the neighboring key points into vectors; for trajectories, we can just sample key points with a fixed temporal interval (0.1 second), starting from $t=0$, and connect them into vectors. Given small enough spatial or temporal intervals, the resulting polylines serve as close approximations of the original map and trajectories. 

Our vectorization process is a one-to-one mapping between continuous trajectories, map annotations and the vector set, although the latter is unordered. This allows us to form a graph representation on top of the vector sets, which can be encoded by graph neural networks. More specifically, we treat each vector $\mathbf{v}_i$ belonging to a polyline $\mathcal{P}_j$ as a node in the graph with node features given by
\begin{equation}
    \mathbf{v}_i = \left[\mathbf{d}_i^s, \mathbf{d}_i^e, \mathbf{a}_i, j \right],
\end{equation}
where $\mathbf{d}_i^s$ and $\mathbf{d}_i^e$ are coordinates of the start and end points of the vector, $\mathbf{d}$ itself can be represented as $(x, y)$ for 2D coordinates or $(x, y, z)$ for 3D coordinates; $\mathbf{a}_i$ corresponds to attribute features, such as object type, timestamps for trajectories, or road feature type or speed limit for lanes; $j$ is the integer id of $\mathcal{P}_j$, indicating $\mathbf{v}_i \in \mathcal{P}_j$.

To make the input node features invariant to the locations of target agents, we normalize the coordinates of all vectors to be centered around the location of target agent at its last observed time step. A future work is to share the coordinate centers for all interacting agents, such that their trajectories can be predicted in parallel.

%\chen{Do we assume the attributes for lanes, regions and trajectories share the same space or not? If they do not, do we need different graph encoders for different components?}
%\jiyang{In Open Dataset, we actually don't have attributes, so all vector share the same encoder;  in Argo, it only provides lane vectors, which has attributes, and they are in the same space, so also share the same encoder. }

\subsection {Constructing the polyline subgraphs}

\begin{figure}[t]
    \centering
    \includegraphics[width=0.7\linewidth]{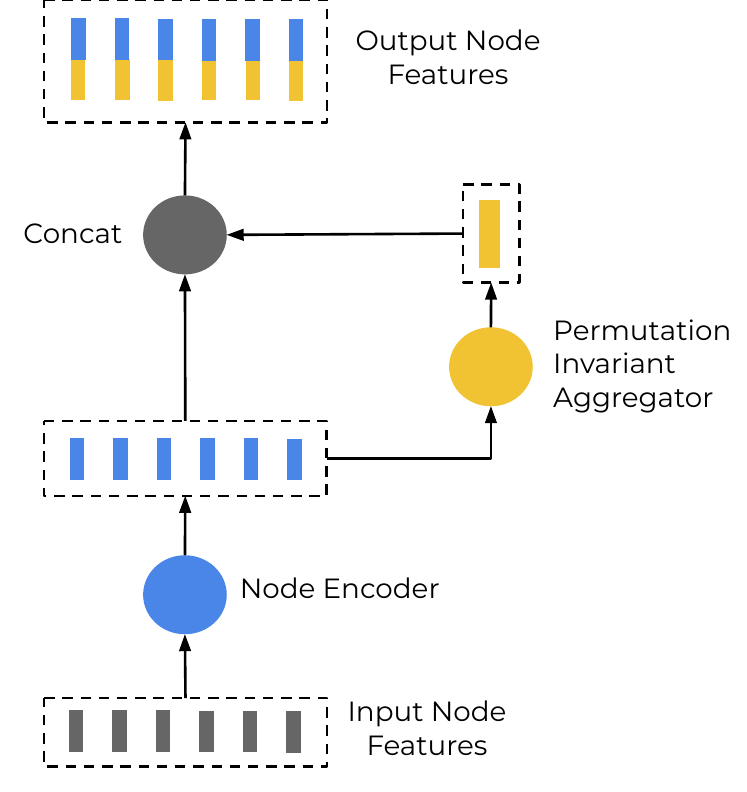}
    \caption{The computation flow on the vector nodes of the same polyline.}
    \label{fig:subgraph}
\end{figure}

To exploit the spatial and semantic locality of the nodes, we take a hierarchical approach by first constructing subgraphs at the vector level, where all vector nodes belonging to the same polyline are connected with each other. Considering a polyline $\mathcal{P}$ with its nodes $\left\{\mathbf{v}_1, \mathbf{v}_2, ..., \mathbf{v}_P\right\}$, we define a single layer of subgraph propagation operation as
\begin{equation}
    \mathbf{v}_i^{(l+1)} = \varphi_\text{rel}\left(g_\text{enc}(\mathbf{v}_i^{(l)}), \varphi_\text{agg}\left(\left\{g_\text{enc}(\mathbf{v}_j^{(l)})\right\}\right)\right)
\end{equation}
where $\mathbf{v}_i^{(l)}$ is the node feature for $l$-th layer of the subgraph network, and $\mathbf{v}_i^{(0)}$ is the input features $\mathbf{v}_i$. Function $g_\text{enc}(\cdot)$ transforms the individual node features, $\varphi_\text{agg}(\cdot)$ aggregates the information from all neighboring nodes, and $\varphi_\text{rel}(\cdot)$ is the relational operator between node $\mathbf{v}_i$ and its neighbors.

In practice, $g_\text{enc}(\cdot)$ is a multi-layer perceptron (MLP) whose weights are shared over all nodes; specifically, the MLP contains a single fully connected layer followed by layer normalization \cite{ba2016layer} and then ReLU  non-linearity. $\varphi_\text{agg}(\cdot)$ is the maxpooling operation, and $\varphi_\text{rel}(\cdot)$ is a simple concatenation. An illustration is shown in Figure~\ref{fig:subgraph}. We stack multiple layers of the subgraph networks, where the weights for $g_\text{enc}(\cdot)$ are different. Finally, to obtain polyline level features, we compute
\begin{equation}
    \mathbf{p} = \varphi_\text{agg}\left(\left\{\mathbf{v}_i^{(L_p)}\right\}\right)
\end{equation}
where $\varphi_\text{agg}(\cdot)$ is again maxpooling.

Our polyline subgraph network can be seen as a generalization of PointNet~\cite{qi2017pointnet}: when we set $\mathbf{d}^s = \mathbf{d}^e$ and let $\mathbf{a}$ and $\mathbf{l}$ to be empty, our network has the same inputs and compute flow as PointNet. However, by embedding the ordering information into vectors, constraining the connectivity of subgraphs based on the polyline groupings, and encoding attributes as node features, our method is particularly suitable to encode structured map annotations and agent trajectories.

\subsection{Global graph for high-order interactions}

%\chen{We need to make text more generic, not focusing too much on transformers. any graph should work here.}

%\chen{We can just mention we construct a global graph with transformers, then we need to explain in detail how to identify each node, and what is the node completion task.}

We now consider modeling the high-order interactions on the polyline node features $\{\mathbf{p}_1, \mathbf{p}_2, ..., \mathbf{p}_P\}$ with a global interaction graph:

\begin{equation}
\left\{\mathbf{p}_i^{(l+1)}\right\} = \text{GNN}\left(\left\{\mathbf{p}_i^{(l)}\right\}, \mathcal{A}\right)
\end{equation}
where $\{\mathbf{p}_i^{(l)}\}$ is the set of polyline node features, $\text{GNN}(\cdot)$ corresponds to a single layer of a graph neural network, and $\mathcal{A}$ corresponds to the adjacency matrix for the set of polyline nodes.

The adjacency matrix $\mathcal{A}$ can be provided a heuristic, such as using the spatial distances~\cite{SocialLSTM} between the nodes. For simplicity, we assume $\mathcal{A}$ to be a fully-connected graph. Our graph network is implemented as a self-attention operation~\cite{vaswani2017attention}:
\begin{equation}
    \text{GNN}(\mathbf{P}) = \text{softmax}\left(\mathbf{P}_Q\mathbf{P}_K^T\right)\mathbf{P}_V
\end{equation}
where $\mathbf{P}$ is the node feature matrix and $\mathbf{P}_Q$, $\mathbf{P}_K$ and $\mathbf{P}_V$ are its linear projections.

We then decode the future trajectories from the nodes corresponding the  moving agents:
\begin{equation}
    \mathbf{v}_i^\text{future} = \varphi_\text{traj}\left(\mathbf{p}_i^{(L_t)}\right)
\end{equation}
where $L_t$ is the number of the total number of GNN layers, and $\varphi_\text{traj}(\cdot)$ is the trajectory decoder. For simplicity, we use an MLP as the decoder function. More advanced decoders, such as the anchor-based approach from MultiPath~\cite{chai2019multipath}, or variational RNNs~\cite{VRNN,sun2019stochastic} can be used to generate diverse trajectories; these decoders are complementary to our input encoder.

We use a single GNN layer in our implementation, so that during inference time, only the node features corresponding to the target agents need to be computed. However, we can also stack multiple layers of $\text{GNN}(\cdot)$ to model higher-order interactions when needed.

To encourage our global interaction graph to better capture interactions among different trajectories and map polylines, we introduce an auxiliary graph completion task. During training time, we randomly mask out the features for a subset of polyline nodes, \eg $\mathbf{p}_i$. We then attempt to recover its masked out feature as:
\begin{equation}
    \mathbf{\hat{p}}_i = \varphi_\text{node}\left(\mathbf{p}_i^{(L_t)}\right)
\end{equation}
where $\varphi_\text{node}(\cdot)$ is the node feature decoder implemented as an MLP. These node feature decoders are not used during inference time.

Recall that $\mathbf{p}_i$ is a node from a fully-connected, unordered graph. In order to identify an individual polyline node when its corresponding feature is masked out, we compute the minimum values of the start coordinates from all of its belonging vectors to obtain the identifier embedding $\mathbf{p}^\text{id}_i$. The inputs node features then become
\begin{equation}
\mathbf{p}_i^{(0)} = \left[\mathbf{p}_i;\mathbf{p}^\text{id}_i\right]
\end{equation}

Our graph completion objective is closely related to the widely successful BERT~\cite{devlin2018bert} method for natural language processing, which predicts missing tokens based on bidirectional context from discrete and sequential text data. We generalize this training objective to work with unordered graphs. Unlike several recent methods (\eg \cite{su2019vl}) that generalizes the BERT objective to unordered image patches with pre-computed visual features, our node features are jointly optimized in an end-to-end framework.

\subsection{Overall framework}
\label{sec:head}

Once the hierarchical graph network is constructed, we optimize for the multi-task training objective

\begin{equation}
    \mathcal{L} = \mathcal{L}_\text{traj} + \alpha \mathcal{L}_\text{node}
\end{equation}

where $\mathcal{L}_\text{traj}$ is the negative Gaussian log-likelihood for the groundtruth future trajectories, $\mathcal{L}_\text{node}$ is the Huber loss between predicted node features and groundtruth masked node features, and $\alpha = 1.0$ is a scalar that balances the two loss terms. To avoid trivial solutions for $\mathcal{L}_\text{node}$ by lowering the magnitude of node features, we L2 normalize the polyline node features before feeding them to the global graph network.

Our predicted trajectories are parameterized as per-step coordinate offsets, starting from the last observed location. We rotate the coordinate system based on the heading of the target vehicle at the last observed location.

%% file: sections/experiments.tex
\section{Experiments}

%\cs{maybe say a little more here}
In this section, we first describe the experimental settings, including the datasets, metrics and  rasterized + ConvNets baseline. Secondly, comprehensive ablation studies are done for both the rasterized baseline and \model. 
%For rasterized representation with ConvNets, we evaluate how rendering resolution and receptive field affect the performance; for our vectorized representation with \model, we conduct various ablation studies to demonstrate its effectiveness and efficiency. 
Thirdly, we compare and discuss the computation cost, including FLOPs and number of parameters. Finally, we compare the performance with state-of-the-art methods.

\subsection{Experimental setup}

\subsubsection{Datasets} 
%
%{\bf In-house dataset} is a  multi-sensory dataset for perception and behavior prediction. It contains lidar, camera, road map data and labels for 1000 segments. Each segment contains roughly 200 frames with length of 20s. Train and validation subsets have 798 and 202 segments respectively; the total number of trajectories sum up to 2.2M and 0.55M for train and validation set. There are many roadgraph features in the maps. We use lane boundaries, stop/yield signs, crosswalks and speed bumps in the rasterized and vectorized representations.

We report results on two vehicle behavior prediction benchmarks, the recently released Argoverse dataset~\cite{chang2019argoverse} and  our in-house behavior prediction dataset.\\
\noindent{\bf Argoverse motion forecasting~\cite{chang2019argoverse}} is a dataset designed for vehicle behavior prediction with trajectory histories. There are 333K 5-second long sequences split into 211K training, 41K validation and 80K testing sequences. The creators curated this dataset by mining interesting and diverse scenarios, such as yielding for a merging vehicle, crossing an intersection, etc. The trajectories are sampled at 10Hz, with (0, 2] seconds are used as observation and (2, 5] seconds for trajectory prediction. Each sequence has one ``interesting'' agent whose trajectory is the prediction target. In addition to vehicle trajectories, each sequence is also associated with map information. The future trajectories of the test set are held out. Unless otherwise mentioned, our ablation study reports performance on the validation set. 

%\cs{the test is held out? maybe add this here}
%\cs{ what does core mean?} 
%The core roadgraph features in the dataset are lane centerlines that vehicles should generally follow.

%
\noindent{\bf In-house dataset} is a large-scale dataset collected for behavior prediction. It contains HD map data, bounding boxes and tracks obtained with an automatic in-house perception system, and manually labeled vehicle trajectories. 
%\cs{shouldn't we rather call this test set instead of validation set?}
The total number of vehicle trajectories are 2.2M and 0.55M for train and test sets. Each  trajectory has a length of 4 seconds, where the (0, 1] second is the history trajectory used as observation, and (1, 4] seconds are the target future trajectories to be evaluated. The trajectories are sampled from real world vehicles' behaviors, including stationary, going straight, turning, lane change and reversing, and roughly preserves the natural distribution of driving scenarios. For the HD map features, we include lane boundaries, stop/yield signs, crosswalks and speed bumps.

For both datasets, the input history trajectories are derived from automatic perception systems and are thus noisy. Argoverse's future trajectories are also machine generated, while In-house has manually labeled future trajectories.

\subsubsection{Metrics} 
For evaluation we adopt the widely used Average Displacement Error (ADE) computed over the entire trajectories and the Displacement Error at $t$ (DE@$t$s) metric, where $t\in\{1.0,2.0,3.0\}$ seconds. The displacements are measured in meters.

\subsubsection{Baseline with rasterized images}
We render $N$ consecutive past frames, where $N$ is 10 for the in-house dataset and 20 for the Argoverse dataset. Each frame is a 400$\times$400$\times$3 image, which has road map information and the detected object bounding boxes. 400 pixels correspond to 100 meters in the in-house dataset, and 130 meters in the Argoverse dataset. Rendering is based on the position of self-driving vehicle in the last observed frame; the self-driving vehicle is placed at the coordinate location (200, 320) in in-house dataset, and (200, 200) in Argoverse dataset. All $N$ frames are stacked together to form a 400$\times$400$\times$3N image as model input.

Our baseline uses a ConvNet to encode the rasterized images, whose architecture is comparable to IntentNet \cite{casas2018intentnet}: we use a ResNet-18 \cite{ResNet16} as the ConvNet backbone. Unlike IntentNet, we do not use the LiDAR inputs. 
%
%\cs{how do you combine the features vectors, how is the spatial information preserved?}
To obtain vehicle-centric features, we crop the feature patch around the target vehicle from the convolutional feature map, and average pool over all the spatial locations of the cropped feature map to get a single vehicle feature vector. We empirically observe that using a deeper ResNet model or rotating the cropped features based on target vehicle headings do not lead to better performance. The vehicle features are then fed into a fully connected layer (as used by IntentNet) to predict the future coordinates in parallel. The model is optimized on 8 GPUs with synchronous training. We use the Adam optimizer~\cite{kingma2014adam} and decay the learning rate every 5 epochs by a factor of 0.3. We train the model for a total of 25 epochs with an initial learning rate of 0.001.

To test how convolutional receptive fields and feature cropping strategies influence the performance, we conduct ablation study on the network receptive field, feature cropping strategy and input image resolutions.

%, which is the same prediction head used by VectorNet. 

\eat{
To test how convolutional receptive fields and feature cropping strategies influence the performance, we conduct the following ablation studies: 

\noindent{\bf Network receptive field}: we vary the convolutional kernel size of ResNet-18 from 3, 5 to 7 to study the impact of convolutional receptive field.

\noindent{\bf Agent-centric feature patch}: we include more spatial context information by  increasing the crop size of convolutional features around the final position of target agents from 1, 3 to 5; or cropping the features along the history locations of the target agent. The cropped feature patches are avg-pooled to get the agent feature.

\noindent{\bf Input resolution}: we keep the context region around SDC the same ($100\times100$ meters), but vary the rendered image size among $100 \times 100$, $200 \times 200$ and $400 \times 400$.}

\begin{table*}[!tb]
\centering
\begin{tabular}{|c|c|c||c|c|c|c|c|c|c|c|}
\hline
Resolution & Kernel & Crop & \multicolumn{4}{c|}{In-house dataset} & \multicolumn{4}{c|}{Argoverse dataset} \\ \cline{4-11} 
                      &      &                       & DE@1s   & DE@2s   & DE@3s   & ADE     & DE@1s    & DE@2s  & DE@3s  & ADE  \\ \hline \hline
100$\times$100 &3$\times$3                       & 1$\times$1                    &  0.63   &  0.94  &   1.32      &     0.82 &    1.14    &     2.80     &     5.19   &    2.21     \\
200$\times$200 &3$\times$3                       & 1$\times$1                     &   0.57   & 0.86  &   1.21      &     0.75   &   1.11     &    2.72      &   4.96 &  2.15       \\ \hline
400$\times$400 & 3$\times$3                       & 1$\times$1                     & 0.55    & 0.82    & 1.16    & 0.72    &   1.12     &    2.72      & 4.94      &   2.16    \\
400$\times$400 &3$\times$3                       & 3$\times$3                     & 0.50    & 0.77    & 1.09    & 0.68    &    1.09    &      2.62    &    4.81      &   2.08      \\ 
400$\times$400 &3$\times$3                       & 5$\times$5                     & 0.50    & 0.76    & 1.08    & 0.67    &    1.09    &     2.60     &    4.70      &  2.08       \\ 
400$\times$400 &3$\times$3                       & traj                     & 0.47    & 0.71    & 1.00    & 0.63    &    1.05    &     2.48     &    4.49      &  1.96       \\ \hline
400$\times$400 &5$\times$5                       & 1$\times$1                     & 0.54    & 0.81    & 1.16    & 0.72    &    1.10    &     2.63     &    4.75      &  2.13   \\ 
400$\times$400 &7$\times$7                       & 1$\times$1                     & 0.53    & 0.81    & 1.16    & 0.72    &     1.10   &    2.63      &    4.74     &    2.13     \\ \hline

\end{tabular}
\caption{Impact of receptive field (as controlled by convolutional kernel size and crop strategy) and rendering resolution for the ConvNet baseline. We report DE and ADE (in meters) on both the in-house dataset and the Argoverse dataset.}\label{tbl:rendered_od}
\end{table*}

\subsubsection{\model with vectorized representations}

To ensure a fair comparison, the vectorized representation takes as input  the same information as the rasterized representation. Specifically, we extract exactly the same set of map features as  when rendering. We also make sure that the visible road feature vectors for a target agent are the same as in the rasterized representation. However, the vectorized representation does enjoy the benefit of  incorporating more complex road features which are non-trivial to render.

Unless otherwise mentioned, we use three graph layers for the polyline subgraphs, and one graph layer for the global interaction graph. The number of hidden units in all MLPs are fixed to 64. The MLPs are followed by layer normalization and ReLU nonlinearity. We normalize the vector coordinates to be centered around the location of target vehicle at the last observed time step.
Similar to the rasterized model, \model is trained on 8 GPUs synchronously with Adam optimizer. The learning rate is decayed every 5 epochs by a factor of 0.3, we train the model for a total of 25 epochs with initial learning rate of 0.001.

To understand the impact of the components on the performance of \model, we conduct ablation studies on the type of context information, \ie whether to use only map or also the trajectories of other agents as well as the impact of number of graph layers for the polyline subgraphs and global interaction graphs.

\subsection{Ablation study for the ConvNet baseline} 

We conduct ablation studies on the impact of ConvNet receptive fields, feature cropping strategies, and the resolution of the rasterized images.

\noindent\textbf{Impact of receptive fields.} As behavior prediction often requires capturing long range road context, the convolutional receptive field could be critical to the prediction quality. We evaluate different variants to see how two key factors of receptive fields, convolutional kernel sizes and feature cropping strategies, affect the prediction performance. The results are shown in Table \ref{tbl:rendered_od}. By comparing kernel size 3, 5 and 7 at 400$\times$400 resolution, we can see that a larger kernel size leads to slight performance improvement. However, it also leads to quadratic increase of the computation cost. We also compare different cropping methods, by increasing the crop size or cropping along the vehicle trajectory at all observed time steps. From the 3rd to 6th rows of Table \ref{tbl:rendered_od} we can see that a larger crop size (3 v.s.\ 1) can significantly improve the performance, and cropping along observed trajectory also leads to better performance. This observation confirms the importance of receptive fields when rasterized images are used as inputs. It also highlights its limitation, where a carefully designed cropping strategy is needed, often at the cost of increased computation cost.

% For the feature cropping method, we can see that cropping larger feature patches (crop size 3 vs 1) can significantly improve the performance. Cropping the feature patch along the history trajectory rather than the final position can further boost the performance (DE@3s) from 1.08 to 1.00 on in-house dataset and 4.70 to 4.49 on ArgoVerse dataset. The performance improvement also shows that the feature receptive field at each location of the ConvNets feature map is limited, it is necessary and critical to enlarge the feature context region for rendering+ConvNets based methods.

\noindent\textbf{Impact of rendering resolution.} We further vary the resolutions of rasterized images to see how it affects the prediction quality and computation cost, as shown in the first three rows of Table \ref{tbl:rendered_od}. We test three different resolutions, including $400 \times 400$ (0.25 meter per pixel), $200 \times 200$ (0.5 meter per pixel) and $100 \times 100$ (1 meter per pixel). It can be seen that the performance increases generally as the resolution goes up. However, for the Argoverse dataset we can see that increasing the resolution from 200$\times$200 to 400$\times$400 leads to slight drop in performance, which can be explained by the decrease of effective receptive field size with the fixed 3$\times$3 kernel.
We discuss the impact on computation cost of these design choices in Section~\ref{sec:flops}.

\begin{table*}[!tb]
\centering
\begin{tabular}{|c|c||c|c|c|c|c|c|c|c|}
\hline
Context & Node Compl. & \multicolumn{4}{c|}{In-house dataset} & \multicolumn{4}{c|}{Argoverse dataset} \\ \cline{3-10} 
                      &                       & DE@1s   & DE@2s   & DE@3s   & ADE     & DE@1s    & DE@2s  & DE@3s  & ADE  \\ \hline \hline
none & - & 0.77 & 0.99 & 1.29 & 0.92&   1.29     &    2.98      &    5.24      &   2.36      \\  \hline
map & no &  0.57 & 0.81  &  1.11  & 0.72   &     0.95   &    2.18      &    3.94      &    1.75     \\
%agents & star &  - &  - & -   & -   &     -   &    -      &    -      &    -     \\
map + agents & no & 0.55 & 0.78 & 1.05 & 0.70  &    0.94    &     2.14     &    3.84      &   1.72      \\ \hline
map & yes & 0.55 & 0.78 & 1.07 & 0.70    &     0.94   &    2.11      &     3.77     &    1.70     \\
%agents & full & 0.58 & 0.80 & 1.09 & 0.72    &   -     &     -     &    -      &   -      \\ 
map + agents & yes  &  {\bf 0.53}     & {\bf 0.74} &  {\bf 1.00}    &   {\bf 0.66}    &   {\bf 0.92}     &      {\bf 2.06}    &      {\bf 3.67}    &   {\bf 1.66}      \\ \hline
%map + agents & 2 $\times$ full  &  -     & {\bf -} &  {\bf -}    &   {\bf -}    &   -     &     -     &     -     &  -       \\ 
%map + agents & 3 $\times$ full  &  -     & {\bf -} &  {\bf -}    &   {\bf -}    &    -    &     -     &    -      &   -      \\ \hline

\end{tabular}
\caption{Ablation studies for \model with different input node types and training objectives. Here ``map'' refers to the input vectors from the HD maps, and ``agents'' refers to the input vectors from the trajectories of non-target vehicles. When ``Node Compl.'' is enabled, the model is trained with the graph completion objective in addition to trajectory prediction. DE and ADE are reported in meters.}\label{tbl:vector_ablation}
\end{table*}

\subsection{Ablation study for \model}

%\chen{There are a few sets of ablations we need to do. First, compare a flat graph with hierarchical graph. Second, compare different ways of modeling global interactions, such as vanilla attention, vanilla GCN and Transformers. Also compare with different depth, width and multiple heads. Third, verify that the auxiliary task actually helps.}

\noindent\textbf{Impact of input node types.} We study whether it is helpful to incorporate both map features and agent trajectories for \model. The first three rows in Table~\ref{tbl:vector_ablation} correspond to using only the past trajectory of the target vehicle (``none'' context), adding only map polylines (``map''), and finally adding trajectory polylines (``map + agents''). We can clearly observe that adding map information significantly improves the trajectory prediction performance. Incorporating trajectory information furthers improves the performance.

\noindent\textbf{Impact of node completion loss.} The last four rows of Table~\ref{tbl:vector_ablation} compares the impact of adding the node completion auxiliary objective. We can see that adding this objective consistently helps with performance, especially at longer time horizons.

\begin{table}[!tb]
\small
\centering
\begin{tabular}{|c|c|c|c|c|c|}\hline
\multicolumn{2}{|c|}{Polyline Subgraph} & \multicolumn{2}{c|}{Global Graph} & \multicolumn{2}{c|}{DE@3s} \\
\hline
Depth & Width & Depth & Width & In-house & Argoverse \\
\hline\hline 

1 & 64 & 1 & 64 & 1.09 & 3.89 \\
3 & 64 & 1 & 64 & 1.00 & 3.67 \\
3 & 128 & 1 & 64 & 1.00 & 3.93 \\
3 & 64 & 2 & 64 & 0.99 & 3.69 \\
3 & 64 & 2 & 256 & 1.02 & 3.69 \\\hline
\end{tabular}
\caption{Ablation on the depth and width of polyline subgraph and global graph. The depth of polyline subgraph has biggest impact on DE@3s.}\label{tbl:vectornet_arch}
\end{table}

\begin{figure}
    \centering
    \includegraphics[width=0.99\linewidth]{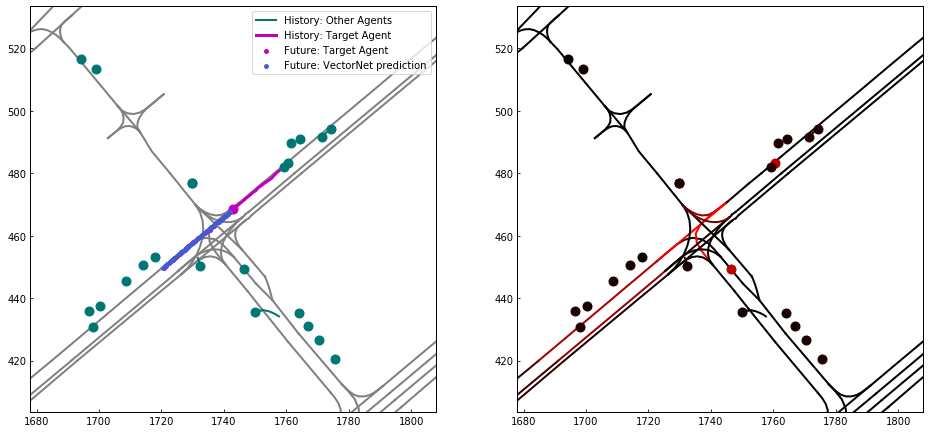}\hfill
    \includegraphics[width=0.99\linewidth]{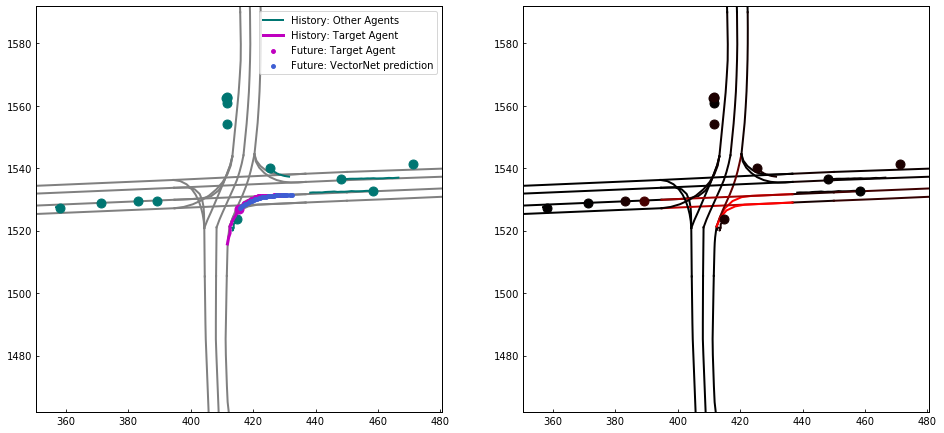}\vspace{2pt}
    \includegraphics[width=0.99\linewidth]{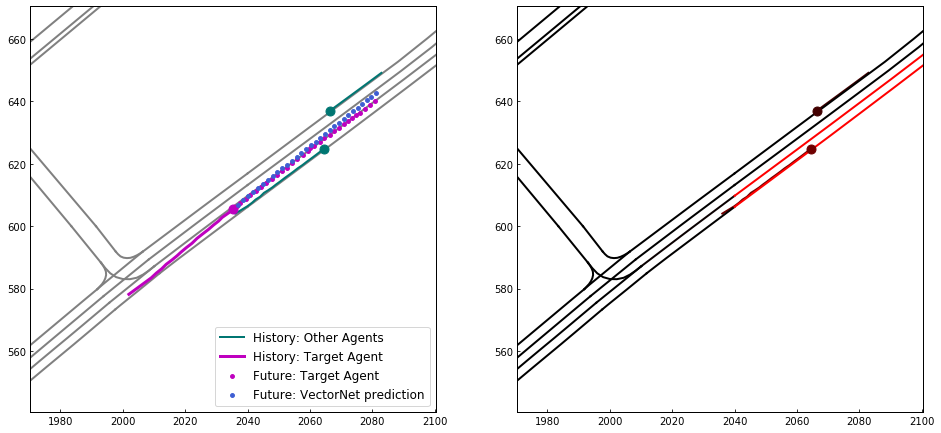}\hfill
    \includegraphics[width=0.99\linewidth]{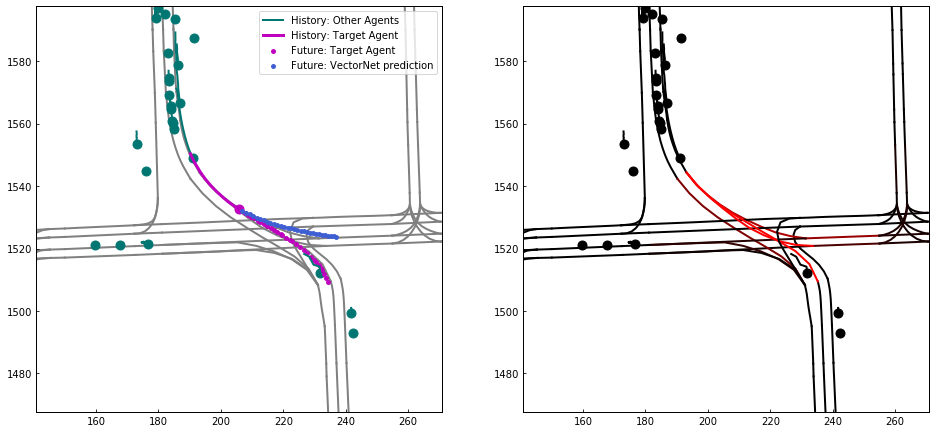}
    \caption{{\it(Left)} Visualization of the prediction: lanes are shown in grey, non-target agents are green, target agent's ground truth trajectory is in pink, predicted trajectory in blue. {\it(Right)} Visualization of attention for road and agent: Brighter \textcolor{red}{red} color corresponds to higher attention score. It can be seen that when agents are facing multiple choices (first two examples), the attention mechanism is able to focus on the correct choices (two right-turn lanes in the second example). The third example is a lane-changing agent, the attended lanes are the current lane and target lane. In the fourth example, though the prediction is not accurate, the attention still produces a reasonable score on the correct lane.}\label{fig:viz}
\end{figure}

\noindent\textbf{Impact on the graph architectures.} In Table~\ref{tbl:vectornet_arch} we study the impact of depths and widths of the graph layers on trajectory prediction performance. We observe that for the polyline subgraph three layers gives the best performance, and for the global graph just one layer is needed. Making the MLPs wider does not lead to better performance, and hurts for Argoverse, presumably because it has a smaller training dataset. Some example visualizations on predicted trajectory and lane attention are shown in Figure~\ref{fig:viz}.

\noindent\textbf{Comparison with ConvNets.} Finally, we compare our \model with the best ConvNet model in Table~\ref{tbl:flops}. For the in-house dataset, our model achieves on par performance with the best ResNet model, while being much more economically in terms of model size and FLOPs. For the Argoverse dataset, our approach significantly outperforms the best ConvNet model with 12\% reduction in DE@3.
We observe that the in-house dataset contains a lot of stationary vehicles due to its natural distribution of driving scenarios; those cases can be easily solved by ConvNets, which are good at capturing local pattern. However, for the Argoverse dataset where only ``interesting'' cases are preserved, \model outperforms the best ConvNet baseline by a large margin; presumably due to its ability to capture long range context information via the hierarchical graph network. 
%\begin{figure}
%    \centering
%    \includegraphics[width=0.9\linewidth]{argo1.png}\hfill
%    \includegraphics[width=0.9\linewidth]{argo2.png}\vspace{2pt}
%    \includegraphics[width=0.9\linewidth]{argo3.png}\hfill
%    \includegraphics[width=0.9\linewidth]{argo4.png}
%    \caption{Qualitative comparison of trajectory prediction results. Red dots are ground truth trajectories of the target agents; blue dots are predictions from \model; green dots are predictions from baseline rendering-based model.}\label{fig:viz}
%\end{figure}

\begin{table}[!tb]
\small
\centering
\begin{tabular}{|c||c|c|c|c|}\hline
Model       & FLOPs      & \#Param & \multicolumn{2}{c|}{DE@3s}  \\ \cline{4-5}
   &     &  & In-house & Argo  \\ \hline\hline 
R18-k3-c1-r100  &     0.66G   & 246K  & 1.32 & 5.19\\
R18-k3-c1-r200  &     2.64G   & 246K &  1.21 & 4.95 \\
R18-k3-c1-r400  &     10.56G  & 246K  &1.16 &  4.96\\
R18-k5-c1-r400  &     15.81G   & 509K & 1.16 & 4.75\\
R18-k7-c1-r400  &     23.67G   & 902K & 1.16 & 4.74\\
R18-k3-c3-r400  &     10.56G   & 246K & 1.09 & 4.81 \\
R18-k3-c5-r400  &     10.56G   & 246K & 1.08 & 4.70\\ 
R18-k3-t-r400  &   10.56G   & 246K & 1.00 & 4.49\\  \hline
\model w/o aux.  &  0.041G$\times$n     & 72K  & 1.05 & 3.84 \\
\model w aux.  &  0.041G$\times$n     & 72K & \textbf{1.00} & \textbf{3.67} \\\hline

\end{tabular}
\caption{Model FLOPs and number of parameters comparison for ResNet and VectorNet. R18-k$M$-c$N$-r$S$ stands for the ResNet-18 model with kernel size  $M \times M$, crop patch size $N \times N$ and input resolution $S \times S$. Prediction decoder is not counted for FLOPs and parameters.}\label{tbl:flops}
\end{table}

\subsection{Comparison of FLOPs and model size} 
\label{sec:flops}
We now compare the FLOPs and model size between ConvNets and \model, and their implications on performance. The results are shown in Table \ref{tbl:flops}. The prediction decoder is not counted for FLOPs and number of parameters. We can see that the FLOPs of ConvNets increase quadratically with the kernel size and input image size; the number of parameters increases quadratically with the kernel size. As we render the images centered at the self driving vehicle, the feature map can be reused among multiple targets, so the FLOPs of the backbone part is a constant number. However, if the rendered images are target-centered, the FLOPs increases linearly with the number of targets. For \model, the FLOPs depends on the number of vector nodes and polylines in the scene. For the in-house dataset, the average number of road map polylines is 17 containing 205 vectors; the average number of road agent polylines is 59 containing 590 vectors. We calculate the FLOPs based on these average numbers. Note that, as we need to re-normalize the vector coordinates and re-compute the VectorNet features for each target, the FLOPs increase linearly with the number of predicting targets ($n$ in Table \ref{tbl:flops}). 

Comparing R18-k3-t-r400 (the best model among ConvNets) with \model, \model significantly outperforms ConvNets. For computation, ConvNets consumes 200+ times more FLOPs than \model (10.56G vs 0.041G) for a single agent; considering that the average number of vehicles in a scene is around 30 (counted from the in-house dataset), the actual computation consumption of VectorNet is still much smaller than that of ConvNets. At the same time, VectorNet needs 29\% of the parameters of ConvNets (72K vs 246K). Based on the comparison, we can see that VectorNet can significantly boost the performance while at the same time dramatically reducing computation cost.

\subsection{Comparison with state-of-the-art methods}

Finally, we compare VectorNet with several baseline approaches~\cite{chang2019argoverse}  and some state-of-the-art methods on the Argoverse~\cite{chang2019argoverse} test set. We report K=1 results (the most likely predictions) in Table~\ref{tbl:rendered_argoverse}. 
The baseline approaches include the constant velocity baseline, nearest neighbor retrieval, and LSTM encoder-decoder. The state-of-the-art approaches are the winners of Argoverse Forecasting Challenge. %We can see that our best ConvNet model already outperforms the previous state-of-the-art by large margin, which confirms the competitiveness of this baseline. 
It can be seen that \model improves the state-of-the-art performance from 4.17 to 4.01 for the DE@3s metric when K=1.
%Note that the results in Table \ref{tbl:rendered_argoverse} are not directly comparable with the results in Table \ref{tbl:flops} and Table \ref{tbl:vector_ablation}, as the the test examples of the latter are generated in the region (130 meters by 130 meters) centered at SDC to have fair comparison with rendering baseline.

%\chen{add leaderboard result}
\begin{table}[!tb]
\centering
\begin{tabular}{c||cc}\hline
Model    & DE@3s          & ADE          \\ \hline\hline
Constant Velocity~\cite{chang2019argoverse}  &  7.89 & 3.53     \\\
Nearest Neighbor~\cite{chang2019argoverse} &  7.88   &   3.45      \\
LSTM ED~\cite{chang2019argoverse}  &  4.95   &   2.15      \\
Challenge Winner: uulm-mrm & 4.19 & 1.90 \\
Challenge Winner: Jean & 4.17 & 1.86 \\\hline
VectorNet  &   \textbf{4.01}  &  \textbf{1.81}       \\\hline
\end{tabular}
\caption{Trajectory prediction performance on the Argoverse Forecasting test set when number of sampled trajectories K=1. Results were retrieved from the Argoverse leaderboard~\cite{argo_leaderboard} on 03/18/2020.}\label{tbl:rendered_argoverse}
\end{table}

%% file: sections/conclusion.tex
\section{Conclusion and future work}

We proposed to represent the HD map and agent dynamics with a vectorized representation. We designed a novel hierarchical graph network, where the first level aggregates information among vectors inside a polyline, and the second level models the higher-order relationships among polylines. Experiments on the large scale in-house dataset and the public available Argoverse dataset show that the proposed VectorNet outperforms the ConvNet counterpart while at the same time reducing the computational cost by a large margin. VectorNet also achieves state-of-the-art performance (DE@3s, K=1) on the Argoverse test set. A natural next step is to incorporate the VectorNet encoder with a multi-modal trajectory decoder (e.g.~\cite{chai2019multipath,tang_multifuture}) to generate diverse future trajectories.

\vspace{0.1in}

\noindent\textbf{Acknowledgement.} We want to thank Benjamin Sapp and Yuning Chai for their helpful comments on the paper.

%% file: main.bbl
\begin{thebibliography}{10}\itemsep=-1pt

\bibitem{argo_leaderboard}
{\em Argoverse Motion Forecasting Competition}, 2019.
\newblock \\ \footnotesize{\url{
  https://evalai.cloudcv.org/web/challenges/challenge-page/454/leaderboard/1279}}.

\bibitem{SocialLSTM}
Alexandre Alahi, Kratarth Goel, Vignesh Ramanathan, Alexandre Robicquet, Li
  Fei-Fei, and Silvio Savarese.
\newblock {Social LSTM: Human Trajectory Prediction in Crowded Spaces}.
\newblock In {\em CVPR}, 2016.

\bibitem{ba2016layer}
Jimmy~Lei Ba, Jamie~Ryan Kiros, and Geoffrey~E Hinton.
\newblock Layer normalization.
\newblock {\em arXiv preprint arXiv:1607.06450}, 2016.

\bibitem{Battaglia2018arxiv}
Peter~W Battaglia, Jessica~B Hamrick, Victor Bapst, Alvaro Sanchez-Gonzalez,
  Vinicius Zambaldi, Mateusz Malinowski, Andrea Tacchetti, David Raposo, Adam
  Santoro, Ryan Faulkner, Caglar Gulcehre, Francis Song, Andrew Ballard, Justin
  Gilmer, George Dahl, Ashish Vaswani, Kelsey Allen, Charles Nash, Victoria
  Langston, Chris Dyer, Nicolas Heess, Daan Wierstra, Pushmeet Kohli, Matt
  Botvinick, Oriol Vinyals, Yujia Li, and Razvan Pascanu.
\newblock Relational inductive biases, deep learning, and graph networks.
\newblock {\em arXiv preprint arXiv:1806.01261}, 2018.

\bibitem{casas2018intentnet}
Sergio Casas, Wenjie Luo, and Raquel Urtasun.
\newblock Intentnet: Learning to predict intention from raw sensor data.
\newblock In {\em CoRL}, 2018.

\bibitem{chai2019multipath}
Yuning Chai, Benjamin Sapp, Mayank Bansal, and Dragomir Anguelov.
\newblock Multipath: Multiple probabilistic anchor trajectory hypotheses for
  behavior prediction.
\newblock In {\em CoRL}, 2019.

\bibitem{chang2019argoverse}
Ming-Fang Chang, John Lambert, Patsorn Sangkloy, Jagjeet Singh, Slawomir Bak,
  Andrew Hartnett, De Wang, Peter Carr, Simon Lucey, Deva Ramanan, et~al.
\newblock Argoverse: {3D} tracking and forecasting with rich maps.
\newblock In {\em CVPR}, 2019.

\bibitem{VRNN}
Junyoung Chung, Kyle Kastner, Laurent Dinh, Kratarth Goel, Aaron~C Courville,
  and Yoshua Bengio.
\newblock A recurrent latent variable model for sequential data.
\newblock In {\em NeurIPS}, 2015.

\bibitem{NGSIM}
James Colyar and Halkias John.
\newblock Us highway 101 dataset.
\newblock {\em FHWA-HRT-07-030}, 2007.

\bibitem{cui2019multimodal}
Henggang Cui, Vladan Radosavljevic, Fang-Chieh Chou, Tsung-Han Lin, Thi Nguyen,
  Tzu-Kuo Huang, Jeff Schneider, and Nemanja Djuric.
\newblock Multimodal trajectory predictions for autonomous driving using deep
  convolutional networks.
\newblock In {\em ICRA}, 2019.

\bibitem{devlin2018bert}
Jacob Devlin, Ming-Wei Chang, Kenton Lee, and Kristina Toutanova.
\newblock {BERT}: Pre-training of deep bidirectional transformers for language
  understanding.
\newblock {\em arXiv preprint arXiv:1810.04805}, 2018.

\bibitem{felsen17}
Panna Felsen, Pulkit Agrawal, and Jitendra Malik.
\newblock What will happen next? forecasting player moves in sports videos.
\newblock In {\em {ICCV}}, 2017.

\bibitem{SocialGAN}
Agrim Gupta, Justin Johnson, Li Fei-Fei, Silvio Savarese, and Alexandre Alahi.
\newblock Social {GAN}: Socially acceptable trajectories with generative
  adversarial networks.
\newblock In {\em CVPR}, 2018.

\bibitem{ResNet16}
Kaiming He, Xiangyu Zhang, Shaoqing Ren, and Jian Sun.
\newblock Deep residual learning for image recognition.
\newblock In {\em {CVPR}}, 2016.

\bibitem{hong2019rules}
Joey Hong, Benjamin Sapp, and James Philbin.
\newblock Rules of the road: Predicting driving behavior with a convolutional
  model of semantic interactions.
\newblock In {\em CVPR}, 2019.

\bibitem{VAIN}
Yedid Hoshen.
\newblock {VAIN}: Attentional multi-agent predictive modeling.
\newblock {\em arXiv preprint arXiv:1706.06122}, 2017.

\bibitem{kingma2014adam}
Diederik~P Kingma and Jimmy Ba.
\newblock Adam: A method for stochastic optimization.
\newblock {\em arXiv preprint arXiv:1412.6980}, 2014.

\bibitem{Kipf2018icml}
Thomas Kipf, Ethan Fetaya, Kuan-Chieh Wang, Max Welling, and Richard Zemel.
\newblock Neural relational inference for interacting systems.
\newblock In {\em ICML}, 2018.

\bibitem{highDdataset}
Robert Krajewski, Julian Bock, Laurent Kloeker, and Lutz Eckstein.
\newblock The highd dataset: A drone dataset of naturalistic vehicle
  trajectories on german highways for validation of highly automated driving
  systems.
\newblock In {\em ITSC}, 2018.

\bibitem{qi2017pointnet}
Charles~R Qi, Hao Su, Kaichun Mo, and Leonidas~J Guibas.
\newblock Pointnet: Deep learning on point sets for 3d classification and
  segmentation.
\newblock In {\em CVPR}, 2017.

\bibitem{qi2017pointnet++}
Charles~Ruizhongtai Qi, Li Yi, Hao Su, and Leonidas~J Guibas.
\newblock Pointnet++: Deep hierarchical feature learning on point sets in a
  metric space.
\newblock In {\em NIPS}, 2017.

\bibitem{gpt2}
Alec Radford, Jeff Wu, Rewon Child, David Luan, Dario Amodei, and Ilya
  Sutskever.
\newblock Language models are unsupervised multitask learners.
\newblock 2019.

\bibitem{Rhinehart_2019_ICCV}
Nicholas Rhinehart, Rowan McAllister, Kris Kitani, and Sergey Levine.
\newblock {PRECOG}: Prediction conditioned on goals in visual multi-agent
  settings.
\newblock In {\em ICCV}, 2019.

\bibitem{robicquet2016learning}
Alexandre Robicquet, Amir Sadeghian, Alexandre Alahi, and Silvio Savarese.
\newblock Learning social etiquette: Human trajectory understanding in crowded
  scenes.
\newblock In {\em ECCV}, 2016.

\bibitem{su2019vl}
Weijie Su, Xizhou Zhu, Yue Cao, Bin Li, Lewei Lu, Furu Wei, and Jifeng Dai.
\newblock Vl-bert: Pre-training of generic visual-linguistic representations.
\newblock {\em arXiv preprint arXiv:1908.08530}, 2019.

\bibitem{sun2019stochastic}
Chen Sun, Per Karlsson, Jiajun Wu, Joshua~B Tenenbaum, and Kevin Murphy.
\newblock Stochastic prediction of multi-agent interactions from partial
  observations.
\newblock In {\em ICLR}, 2019.

\bibitem{Sun_2019_arXiv}
Chen Sun, Austin Myers, Carl Vondrick, Kevin Murphy, and Cordelia Schmid.
\newblock {VideoBERT}: A joint model for video and language representation
  learning.
\newblock In {\em ICCV}, 2019.

\bibitem{RAF}
Chen Sun, Abhinav Shrivastava, Carl Vondrick, Rahul Sukthankar, Kevin Murphy,
  and Cordelia Schmid.
\newblock Relational action forecasting.
\newblock In {\em CVPR}, 2019.

\bibitem{tang_multifuture}
Charlie Tang and Russ~R Salakhutdinov.
\newblock Multiple futures prediction.
\newblock In {\em NeurIPS}. 2019.

\bibitem{vaswani2017attention}
Ashish Vaswani, Noam Shazeer, Niki Parmar, Jakob Uszkoreit, Llion Jones,
  Aidan~N Gomez, {\L}ukasz Kaiser, and Illia Polosukhin.
\newblock Attention is all you need.
\newblock In {\em NIPS}, 2017.

\bibitem{GAT}
Petar Veli{\v c}kovi{\'c}, Guillem Cucurull, Arantxa Casanova, Adriana Romero,
  Pietro Li{\`o}, and Yoshua Bengio.
\newblock Graph attention networks.
\newblock In {\em ICLR}, 2018.

\bibitem{Yeh_2019_CVPR}
Raymond~A. Yeh, Alexander~G. Schwing, Jonathan Huang, and Kevin Murphy.
\newblock Diverse generation for multi-agent sports games.
\newblock In {\em CVPR}, 2019.

\bibitem{ZhanBasketball}
Eric Zhan, Stephan Zheng, Yisong Yue, Long Sha, and Patrick Lucey.
\newblock Generative multi-agent behavioral cloning.
\newblock {\em arXiv:1803.07612}, 2018.

\end{thebibliography}
